\documentclass{article}
\usepackage{ijcai22}

\usepackage{times}
\usepackage{soul}
\usepackage{url}
\usepackage[hidelinks]{hyperref}
\usepackage[utf8]{inputenc}
\usepackage[small]{caption}
\usepackage{graphicx}
\usepackage{amsmath}
\usepackage{amsthm}
\usepackage{booktabs}
\usepackage{algorithm}
\usepackage{algorithmic}
\usepackage{subfigure}
\usepackage{graphbox}
\usepackage{multirow}

\usepackage[normalem]{ulem}
\useunder{\uline}{\ul}{}
\title{SVTR: Scene Text Recognition with a Single Visual Model}

\author{
Yongkun Du$^{1}$\and
Zhineng Chen$^{2}$\footnote{Corresponding Author}\and
Caiyan Jia$^{1}$\and
Xiaoting Yin$^3$\and
Tianlun Zheng$^2$\and\\
Chenxia Li$^3$\and
Yuning Du$^3$\and
Yu-Gang Jiang$^2$
\affiliations
$^1$School of Computer and Information Technology and Beijing Key Lab of Traffic Data Analysis and Mining, Beijing Jiaotong University, China\\
$^2$Shanghai Collaborative Innovation Center of Intelligent Visual Computing, School of Computer Science, Fudan University, China\\
$^3$Baidu Inc., China
\emails
\{yongkundu,cyjia\}@bjtu.edu.cn,
\{zhinchen,ygj\}@fudan.edu.cn,
tlzheng21@m.fudan.edu.cn,
\{yinxiaoting,lichenxia,duyuning\}@baidu.com
}
\begin{document}

\maketitle

\begin{abstract}
Dominant scene text recognition models commonly contain two building blocks, a visual model for feature extraction and a sequence model for text transcription. This hybrid architecture, although accurate, is complex and less efficient. In this study, we propose a Single Visual model for Scene Text recognition within the patch-wise image tokenization framework, which dispenses with the sequential modeling entirely. The method, termed SVTR, firstly decomposes an image text into small patches named character components. Afterward, hierarchical stages are recurrently carried out by component-level mixing, merging and/or combining. Global and local mixing blocks are devised to perceive the inter-character and intra-character patterns, leading to a multi-grained character component perception. Thus, characters are recognized by a simple linear prediction. Experimental results on both English and Chinese scene text recognition tasks demonstrate the effectiveness of SVTR. SVTR-L (Large) achieves highly competitive accuracy in English and outperforms existing methods by a large margin in Chinese, while running faster. In addition, SVTR-T (Tiny) is an effective and much smaller model, which shows appealing speed at inference. The code is publicly available at \url{https://github.com/PaddlePaddle/PaddleOCR}.

\end{abstract}

\section{Introduction}
Scene text recognition aims to transcript a text in natural image to digital character sequence, which conveys high-level semantics vital for scene understanding. The task is challenging due to variations in text deformations, fonts, occlusions, cluttered background, etc. In the past years, many efforts have been made to improve the recognition accuracy. Modern text recognizers, besides accuracy, also take factors like inference speed into account because of practical requirements.

\begin{figure}[ht] 
\centering
\includegraphics[width=0.4\textwidth]{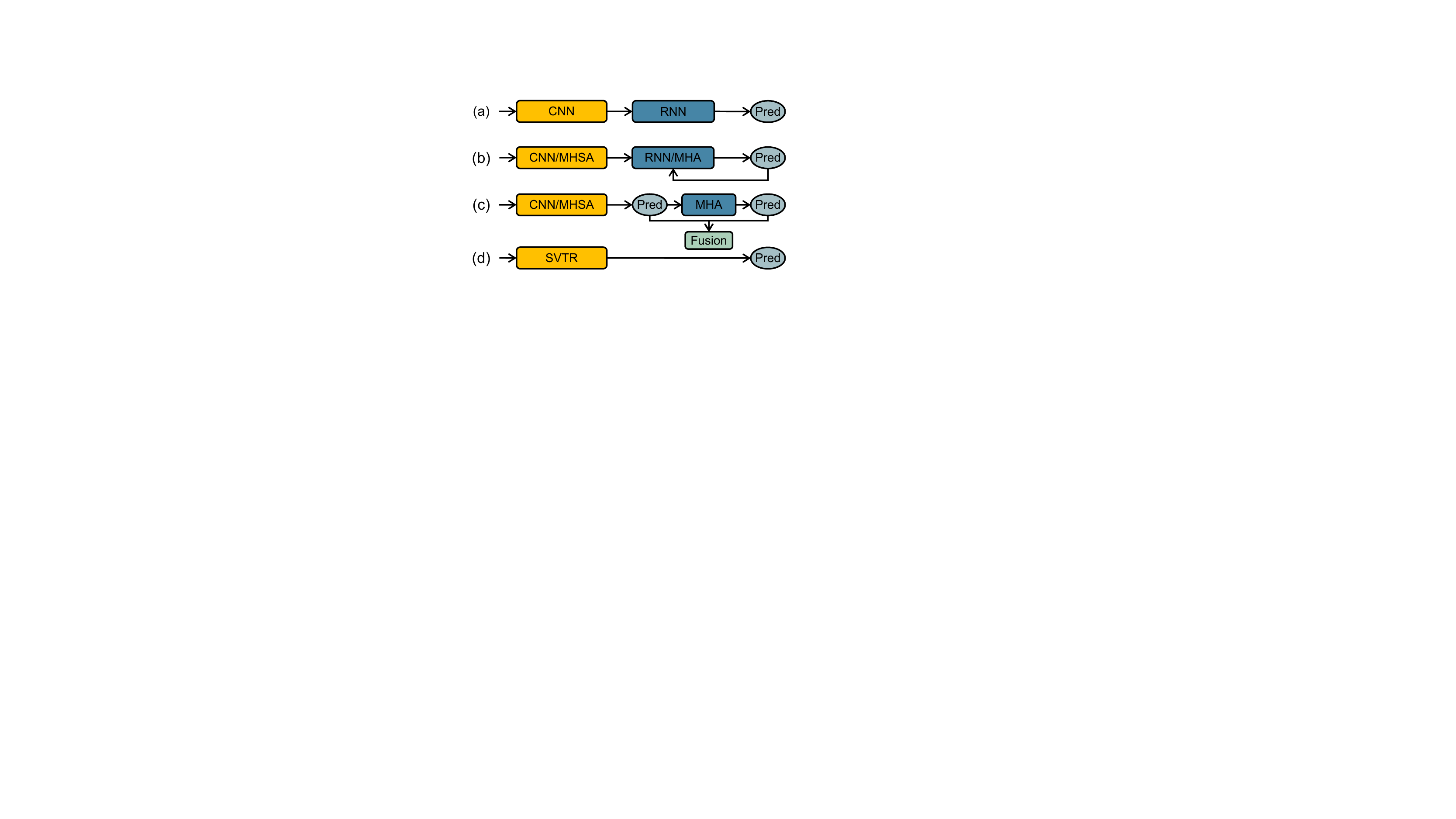}  
\caption{(a) CNN-RNN based models. (b) Encoder-Decoder models. MHSA and MHA denote multi-head self-attention and multi-head attention, respectively. (c) Vision-Language models. (d) Our SVTR, which recognizes scene text with a single visual model and enjoys efficient, accurate and cross-lingual versatile.}  
\label{fig:1}  
\end{figure}


Methodologically, scene text recognition can be viewed as a cross-modal mapping from image to character sequence. Typically, the recognizer consists of two building blocks, a visual model for feature extraction and a sequence model for text transcription. For example, CNN-RNN based models \cite{Zhai2016Chpr,shi2017crnn} first employed CNN for feature extraction. The feature was then reshaped as a sequence and modeled by BiLSTM and CTC loss to get the prediction (Figure \ref{fig:1}(a)). They are featured by efficiency and remain the choice for some commercial recognizers. However, the reshaping is sensitive to text disturbances such as deformation, occlusion, etc, limiting their effectiveness.

\begin{figure*}[ht]  
\centering  
\includegraphics[width=0.85\textwidth]{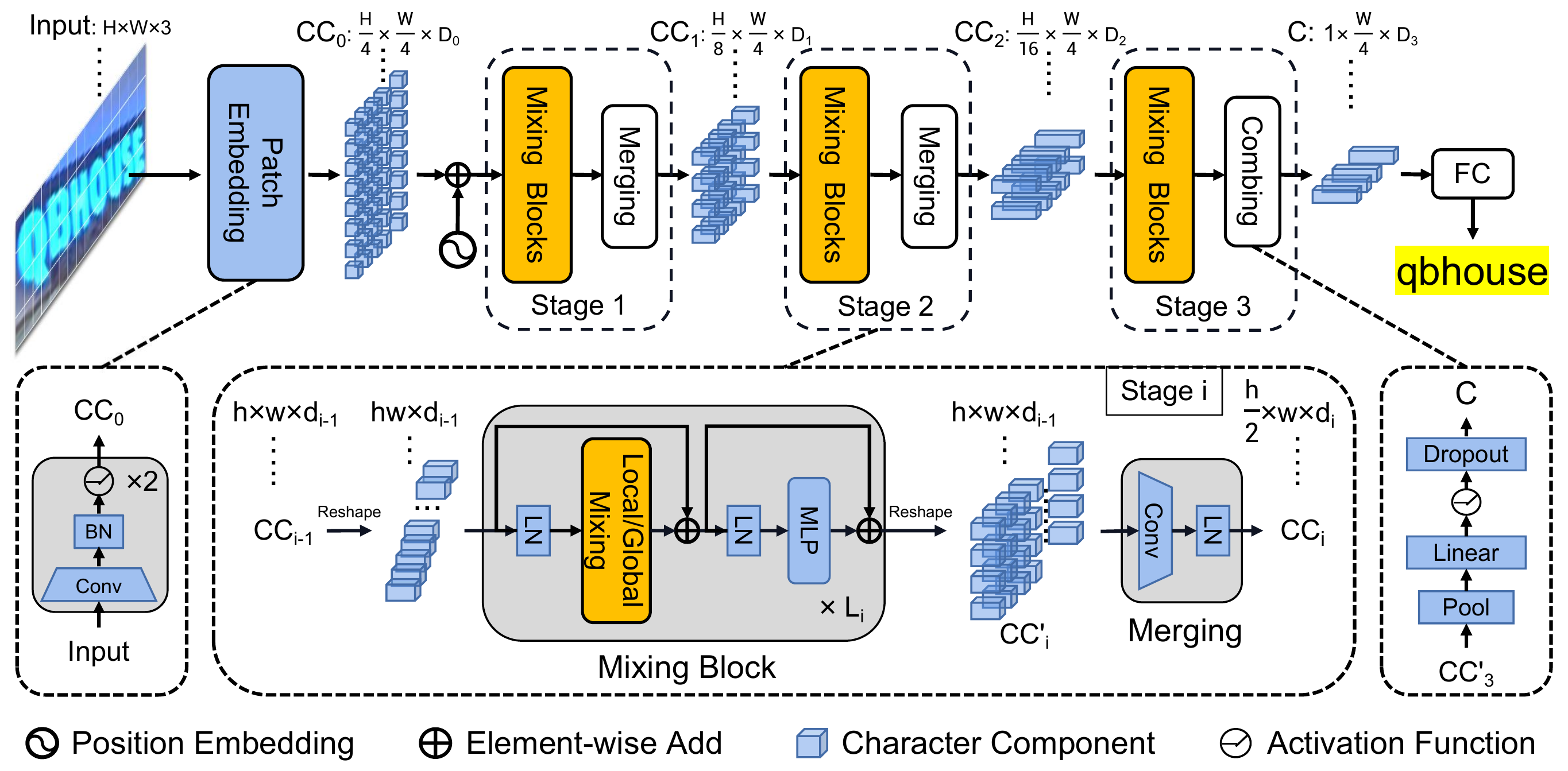}  
\caption{Overall architecture of the proposed SVTR. It is a three-stage height progressively decreased network. In each stage, a series of mixing blocks are carried out and followed by a merging or combining operation. At last, the recognition is conducted by a linear prediction.}  
\label{fig:2}  
\end{figure*}

Later, encoder-decoder based auto-regressive methods became popular \cite{Sheng2019nrtr,li2019sar,zheng2021cdistnet}, the methods transform the recognition as an iterative decoding procedure (Figure \ref{fig:1}(b)).
As a result, improved accuracy was obtained as the context information was considered. However, the inference speed is slow due to the character-by-character transcription. The pipeline was further extended to vision-language based framework \cite{yu2020srn,fang2021abinet}, where language knowledge was incorporated (Figure \ref{fig:1}(c)) and parallel prediction was conducted. However, the pipeline often requires a large capacity model or complex recognition paradigm to ensure the recognition accuracy, restricting its efficiency.



Recently, there are efforts emphasized developing simplified architectures to accelerate the speed. For example, using complex training paradigm but simple model for inference. CRNN-RNN based solution was revisited in \cite{hu2020gtc}. It utilized the attention mechanism and graph neural network to aggregate sequential features corresponding to the same character. At inference, the attention modeling branch was discarded to balance accuracy and speed. PREN2D~\cite{yan2021pren} further simplified the recognition by aggregating and decoding the 1D sub-character features simultaneously. \cite{Wang_2021_visionlan} proposed VisionLAN. It introduced a character-wise occluded learning to endue the visual model with language capability. While at inference, the visual model was applied merely for speedup. In view of the simplicity of a single visual model based architecture, some recognizers were proposed by employing off-the-shelf CNN~\cite{Fedor2018Rosetta} or ViT~\cite{atienza2021vitstr} as the feature extractor. Despite being efficiency, their accuracy is less competitive compared to state-of-the-art methods. 

We argue that the single visual model based scheme is effective only if discriminative character features could be extracted. Specifically, the model could successfully catch both intra-character local patterns and inter-character long-term dependence. The former encodes stroke-like features that describe fine-grained features of a character, being a critical source for distinguishing characters. While the latter records language-analogous knowledge that describes the characters from a complementary aspect. However, the two properties are not well modeled by previous feature extractors. For example, CNN backbones are good at modeling local correlation rather than global dependence. Meanwhile, current transformer-based general-purpose backbones would not give privilege to local character patterns.

Motivated by the issues mentioned above, this work aims to enhance the recognition capability by reinforcing the visual model. To this end, we propose a visual-based model SVTR for accurate, fast and cross-lingual versatile scene text recognition. 
Inspired by the recent success of vision transformer~\cite{dosovitskiy2020vit,liu2021Swin}, SVTR first decomposes an image text into small 2D patches termed character components, as each of which may contain a part of a character only. Thus, patch-wise image tokenization followed by self-attention is applied to capture recognition clues among character components. Specifically, a text-customized architecture is developed for this purpose. It is a three-stage height progressively decreased backbone with mixing, merging and/or combining operations. Local and global mixing blocks are devised and recurrently employed at each stage, together with the merging or combining operation, acquiring the local component-level affinities that represent the stroke-like feature of a character, and long-term dependence among different characters. Therefore, the backbone extracts component features of different distance and at multiple scales, forming a multi-grained character feature perception. As a result, the recognition is reached by a simple linear prediction. In the whole process only one visual model is employed. We construct four architecture variants with varying capacities. Extensive experiments are carried out on both English and Chinese scene text recognition tasks. It is shown that SVTR-L (large) achieves highly competitive results in English and outperforms state-of-the-art models by a large margin in Chinese, while running faster. Meanwhile, SVTR-T (Tiny) is also an effective and much smaller model yet with appealing inference speed. The main contributions of this work can be summarized as follows.

\begin{itemize}
\item We demonstrate, for the first time, that a single visual model can achieve competitive or even higher accuracy as advanced vision-language models in scene text recognition. It is promising to practical applications due to its efficiency and cross-lingual versatility.

\item We propose SVTR, a text-customized model for recognition. It introduces local and global mixing blocks for extracting stroke-like features and inter-character dependence, respectively, together with the multi-scale backbone, forming a multi-grained feature description.

\item Empirical studies on public benchmarks demonstrate the superiority of SVTR. SVTR-L achieves the state-of-the-art performance in recognizing both English and Chinese scene texts. While SVTR-T is effective yet efficient, with parameters of 6.03M and consuming 4.5ms per image text on average in one NVIDIA 1080Ti GPU.
\end{itemize}

\section{Method}

\subsection{Overall Architecture}
Overview of the proposed SVTR is illustrated in Figure \ref{fig:2}. It is a three-stage height progressively decreased network dedicated to text recognition. For an image text of size $H \times  W \times  3$, it is first transformed to $\frac{H}{4} \times  \frac{W}{4}$ patches of dimension $D_0$ via a progressive overlapping patch embedding. The patches are termed character components, each associating with a fraction of text character in the image. Then, three stages, each composed of a series of mixing blocks followed by a merging or combining operation, are carried out at different scales for feature extraction. local and global mixing blocks are devised for stroke-like local pattern extraction and inter-component dependence capturing. With the backbone, component feature and dependence of different distance and at multiple scales are characterized, generating a representation referred to as $C$ of size $1 \times \frac{W}{4} \times D_3$, which perceives multi-grained character features. Finally, a parallel linear prediction with de-duplication is conducted to get the character sequence.

\begin{figure}[t]
	\centering  
	\subfigbottomskip=2pt
	\subfigcapskip=-5pt
	\subfigure[]{
		\includegraphics[ width=0.3\linewidth]{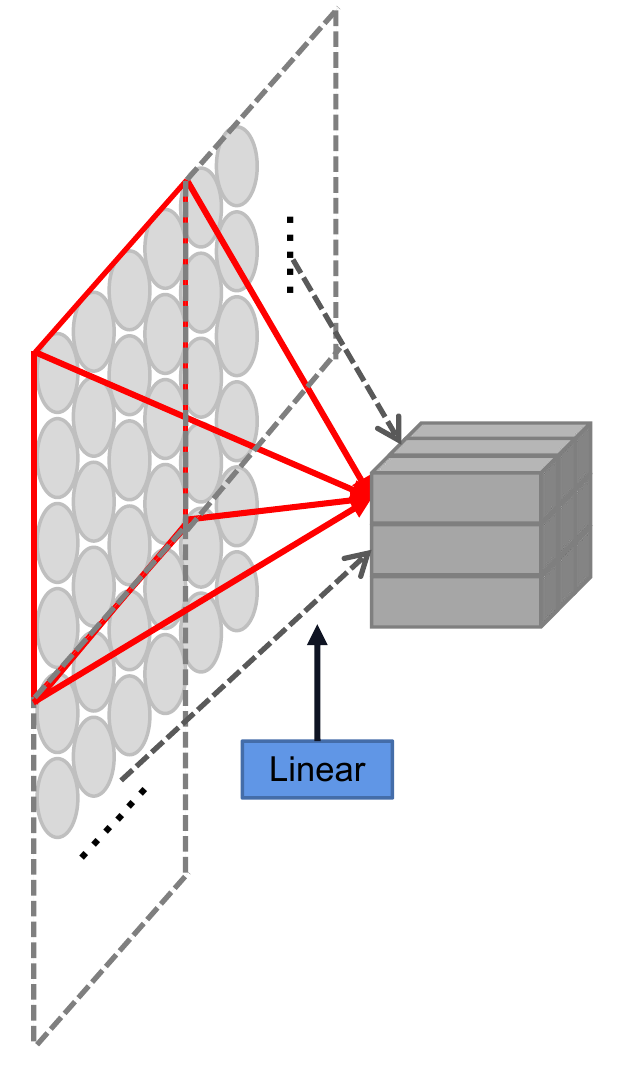}}
	\subfigure[]{
		\raisebox{0.17\height}{\includegraphics[ width=0.45\linewidth]{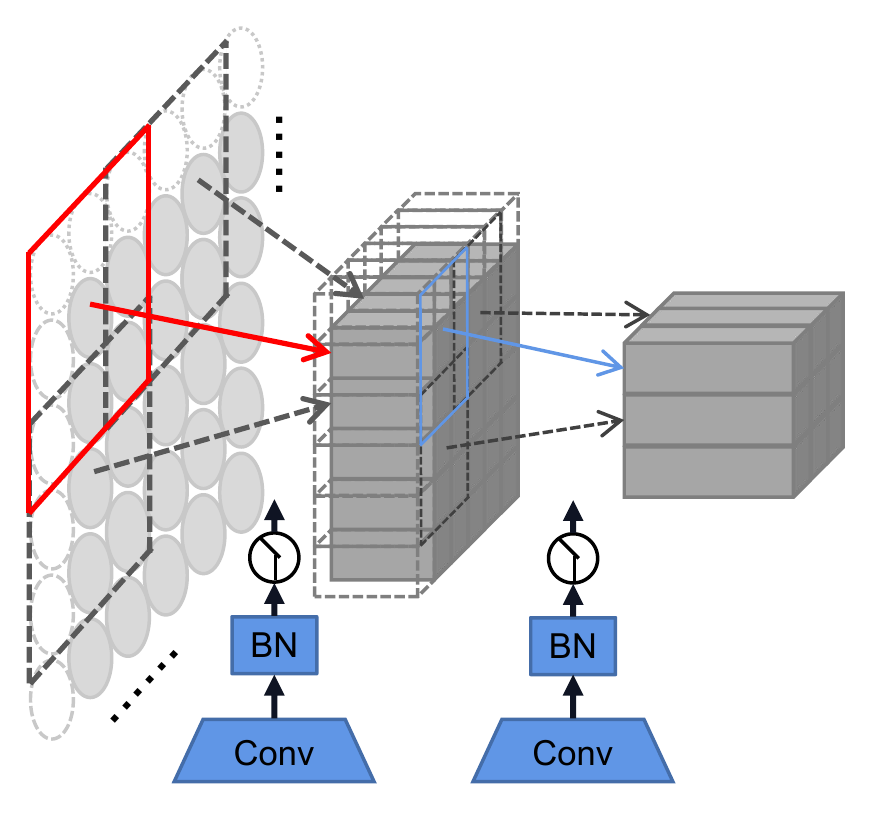}}}
\caption{(a) The linear projection in ViT \protect\cite{dosovitskiy2020vit}. (b) Our progressive overlapping patch embedding.}
\label{fig:3}
\end{figure}
\subsection{Progressive Overlapping Patch Embedding}
With an image text, the first task is to obtain feature patches which represent character components from $X \in  R^{H \times W \times 3}$ to $CC_{0}  \in R^{ \frac{H}{4} \times \frac{W}{4} \times D_{0} }$. 
There exists two common one-step projections for this purpose, i.e., a $4 \times 4$ disjoint linear projection (see Figure \ref{fig:3}(a)) and a $7 \times 7$ convolution with stride 4.
Alternatively, we implement the patch embedding by using two consecutive $3 \times 3$ convolutions with stride 2 and batch normalization, as shown in Figure \ref{fig:3}(b). The scheme, despite increasing the computational cost a little, adds the feature dimension progressively which is in favor of feature fusion. Ablation study in Section \ref{section:3.3} shows its effectiveness.

\subsection{Mixing Block}

Since two characters may differ slightly, text recognition heavily relies on features at character component level. However, existing studies mostly employ a feature sequence to represent the image text. Each feature corresponds to a thin-slice image region, which is often noisy especially for irregular text. It is not optimal for describing the character. Recent advancement of vision transformer introduces 2D feature representation, but how to leverage this representation in the context of text recognition is still worthy investigation.

More specifically, with the embedded components, we argue that text recognition requires two kinds of features. The first is local component patterns such as the stroke-like feature. It encodes the morphology feature and correlation between different parts of a character. The second is inter-character dependence such as the correlation between different characters or between text and non-text components. Therefore, we devise two mixing blocks to perceive the correlation by using self-attention with different reception fields.

\begin{figure}[t]
	\centering  
	\subfigbottomskip=2pt 
	\subfigcapskip=-1pt 
	\subfigure[]{
		\includegraphics[ width=0.4\linewidth]{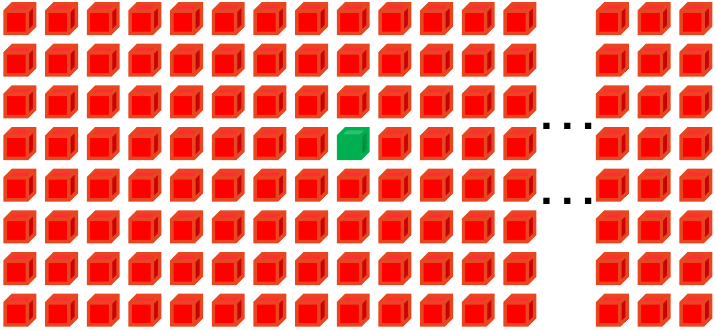}}
	\subfigure[]{
		\includegraphics[ width=0.4\linewidth]{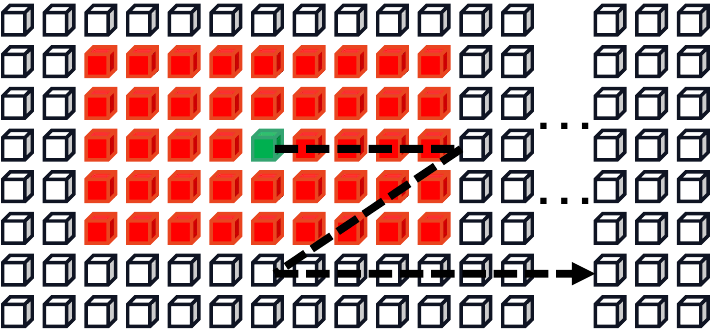}}
\caption{Illustration of (a) global mixing and (b) local mixing.} 
\label{fig:globallocal}  
\end{figure}

\paragraph{Global Mixing.}As seen in Figure \ref{fig:globallocal}(a), global mixing evaluates the dependence among all character components. Since text and non-text are two major elements in an image, such a general-purpose mixing could establish the long-term dependence among component from different characters. Besides, it also capable of weakening the influence of non-text components, while enhancing the importance of text components. Mathematically, for character components $CC_{i-1}$ from the previous stage, it is first reshaped as a feature sequence. When feeding into the mixing block, a layer norm is applied and followed by a multi-head self-attention for dependence modeling. In the following, a layer norm and a MLP are sequentially applied for feature fusion. Together with the shortcut connections, forming the global mixing block.

\paragraph{Local Mixing.}As seen in Figure \ref{fig:globallocal}(b), local mixing evaluates the correlation among components within a predefined window. Its objective is to encode the character morphology feature and establish the associations between components within a character, which simulates the stroke-like feature vital for character identification. Different from global mixing, local mixing considers a neighborhood for each component. Similar to convolution, the mixing is carried out in a sliding window manner. The window size is empirically set to $7 \times 11$. Compared with the global mixing, it implements the self-attention mechanism to capture local patterns.

As aforementioned, the two mixing blocks aims to extract different features that are complementary. In SVTR, the blocks are recurrently applied many times in each stage for comprehensive feature extraction. Permutation of the two kinds of blocks will be ablated later.

\subsection{Merging}

It is computational expensive to maintain a constant spatial resolution 
across stages, which also leads to redundant representation. As a consequence, we devise a merging operation following the mixing blocks at each stage (except the last one). With the features outputted from the last mixing block, we first reshape it to an embedding of size $h \times  w \times  d_{i-1}$, denoting current height, width and channels, respectively. Then, we employ a $3 \times 3$ convolution with stride 2 in the height dimension and 1 in the width dimension, followed by a layer norm, generating an embedding of size $\frac{h}{2} \times  w \times  d_i$.

The merging operation halve the height while keep a constant width. It not only reduce the computational cost, but also build a text-customized hierarchical structure. Typically, most image text appears horizontally or near horizontally. Compressing the height dimension can establish a multi-scale representation for each character, while not affecting the patch layout in the width dimension. Therefore, it would not increase the chance of encoding adjacent characters into the same component across stages. We also increase the channel dimension $d_i$ to compensate the information loss.

\subsection{Combining and Prediction}
In the last stage, the merging operation is replaced by a combining operation. It pools the height dimension to 1 at first, followed by a fully-connected layer, non-linear activation and dropout. By doing this, character components are further compressed to a feature sequence, where each element is represented by a feature of length $D_3$. Compared to the merging operation, the combining operation can avoid applying convolution to an embedding whose size is very small in one dimension, e.g., with 2 in height.

With the combined feature, we implement the recognition by a simple parallel linear prediction. Concretely, a linear classifier with $N$ nodes is employed. It generates a transcript sequence of size $\frac{W}{4}$, where ideally, components of the same character are transcribed to duplicate characters, components of non-text are transcribed to a blank symbol. The sequence is automatically condensed to the final result. In the implementation, $N$ is set to 37 for English and 6625 for Chinese.

\subsection{Architecture Variants}
There are several hyper-parameters in SVTR, including the depth of channel and the number of heads at each stage, the number of mixing blocks and their permutation. By varying them, SVTR architectures with different capacities could be obtained and we construct four typical ones, i.e., SVTR-T (Tiny), SVTR-S (Small), SVTR-B (Base) and SVTR-L (Large). Their detail configurations are shown in Table \ref{tab:booktab1}.

\begin{table*}[t]
\centering
\begin{tabular}{c|c|c|c|c|c|cc}
\toprule
Models & $\left[D_0, D_1, D_2 \right]$ & $[L_1, L_2, L_3]$ & Heads        & $D_3$  & Permutation & Params (M) & FLOPs (G) \\
\midrule
SVTR-T & {[}64,128,256{]}  & {[}3,6,3{]}      & {[}2,4,8{]}  & 192 & $[L]_6 { [G]_6}$  & 4.15   & 0.29  \\
SVTR-S & {[}96,192,256{]}  & {[}3,6,6{]}      & {[}3,6,8{]}  & 192 & $[L]_8  {[G]_7 }$ & 8.45   & 0.63  \\
SVTR-B & {[}128,256,384{]} & {[}3,6,9{]}      & {[}4,8,12{]} & 256 & $[L]_8 { [G]_{10}}$  & 22.66        &  3.55  \\
SVTR-L & {[}192,256,512{]} & {[}3,9,9{]}      & {[}6,8,16{]} & 384 & $[L]_{10} {[G]_{11}}$ & 38.81  & 6.07 \\
\bottomrule
\end{tabular}
\caption{Architecture specifications of SVTR variants (w/o counting the rectification module and linear classifier).}
\label{tab:booktab1}
\end{table*}

\section{Experiments}

\subsection{Datasets}

For English recognition task, our models are trained on two commonly used synthetic scene text datasets, i.e., {\bf MJSynth} (MJ) \cite{jaderberg14synthetic,Jaderberg2015ReadingTI} and {\bf SynthText} (ST) \cite{Synthetic}. Then the models are tested on six public benchmarks as follows: {\bf ICDAR 2013} (IC13) \cite{icdar2013} contains 1095 testing images. we discard images that contain non-alphanumeric characters or less than three characters. As a result, IC13 contains 857 images. {\bf Street View Text} (SVT) \cite{Wang2011SVT} has 647 testing images cropped form Google Street View. Many images are severely corrupted by noise, blur, and low resolution. {\bf IIIT5K-Words} (IIIT) \cite{IIIT5K} is collected from the website and contains 3000 testing images. {\bf ICDAR 2015} (IC15) \cite{icdar2015} is taken with Google Glasses without careful position and focusing. IC15 has two versions with 1,811 images and 2,077 images. We use the former one. {\bf Street View Text-Perspective} (SVTP) \cite{SVTP} is also cropped form Google Street View. There are 639 test images in this set and many of them are perspectively distorted. {\bf CUTE80} (CUTE) is proposed in\cite{Risnumawan2014cute} for curved text recognition. 288 testing images are cropped from full images by using annotated words.

For Chinese recognition task, we use the {\bf Chinese Scene Dataset} \cite{chen2021benchmarking}. It is a public dataset containing 509,164, 63,645 and 63,646 training, validation, and test images. 
The validation set is utilized to determine the best model, which is then assessed using the test set.

\subsection{Implementation Details}

SVTR uses the rectification module~\cite{shi2019aster}, where the image text is resized to $32\times64$ for distortion correction.
We use the AdamW optimizer with weight decay of 0.05 for training. For English models, the initial learning rate are set to $\frac{5}{10^{4}} \times \frac{batchsize}{2048}$. The cosine learning rate scheduler with 2 epochs linear warm-up is used in all 21 epochs. Data augmentation like rotation, perspective distortion, motion blur and Gaussian noise, are randomly performed during the training. The alphabet includes all case-insensitive alphanumerics. The maximum predict length is set to 25. The length exceeds the vast majority of English words. For Chinese models, the initial learning rate are set to $\frac{3}{10^{4}} \times \frac{batchsize}{512}$. The cosine learning rate scheduler with 5 epochs linear warm-up is used in all 100 epochs. Data augmentation was not used for training. The maximum predict length is set to 40. Word accuracy is used as the evaluation metric. All models are trained by using 4 Tesla V100 GPUs on PaddlePaddle.

\subsection{Ablation Study} \label{section:3.3}
To better understand SVTR, we perform controlled experiments on both IC13 (regular text) and IC15 (irregular text) under different settings. For efficiency, all the experiments are carried out by using SVTR-T without rectification module and data augmentation.

\subsubsection{The Effectiveness of Patch Embedding}
As seen in Table \ref{tab:patch_merging} (the left half), different embedding strategies behave slightly different in recognition accuracy. Our progressive embedding scheme outperforms the two default ones by 0.75\% and 2.8\% on average on the two datasets, indicating that it is effective especially for irregular text.

\subsubsection{The Effectiveness of Merging}

There are also two choices, i.e., applying the merging operation to build a progressively decreased network ($prog$), and keeping a constant spatial resolution across stages ($none$). As indicated in the right half of Table \ref{tab:patch_merging}. The merging not only reduces the computational cost, but also increases the accuracy on both datasets. It verifies that a multi-scale sampling on the height dimension is effective for text recognition.

\begin{table}[t]
\centering
\subtable
{ \setlength{\tabcolsep}{3pt}{\begin{tabular}{ccc|ccccc}
      \toprule
Embedding  & IC13 & IC15 & Merging & IC13 & IC15 & FLOPS \\
\midrule
Linear & 92.5 & 72.0 &  None & 92.4 & 71.8 & 1.10 \\
Overlap & 93.0 & 73.9 & Prog &  \textbf{93.5} & \textbf{74.8} & 0.29\\
Ours  & \textbf{93.5} & \textbf{74.8 }
\\
\bottomrule
      \end{tabular}}
}

\caption{Ablation study on patch embedding (the left half) and merging (the right half).}
\label{tab:patch_merging}
\end{table}

\begin{table}[t]
\centering
\setlength{\tabcolsep}{6pt}{
\begin{tabular}{c|cc|c|cc } 
\toprule
            CP   & IC13 & IC15& CP           & IC13 & IC15 \\
           \midrule 
None  & 91.6 & 68.2& ${[}G{]}_{12}$ & 92.7 & 73.7 \\
${[}G{]}_6{[}L{]}_6$ & 92.2 & 71.9 &${[}L{]}_{12}$ & 91.3 & 70.2 \\
${[}L{]}_6{[}G{]}_6$ & \textbf{93.5} & \textbf{74.8} &${[}GL{]}_6$  & 91.9 & 72.4 \\
 & & &$[LG]_6$ & \textbf{93.5} & 73.5
\\
\bottomrule
\end{tabular} }

\caption{Ablation study on mixing block permutation.}
\label{tab:Perception}
\end{table}

\subsubsection{The Permutation of Mixing Blocks}
There are various ways to group the global and local mixing blocks in each stage. Several of them are assessed in Table \ref{tab:Perception}, where $none$ means no mixing block is taken into account. ${[}G{]}_6{[}L{]}_6$ means for each stage, six global mixing blocks are carried out at first, and then six local mixing blocks. Others are defined similarly. As can be seen, almost every scheme gives certain accuracy improvement. We believe that the improvements are attributed to perceiving comprehensive character component features. The relatively large gains on irregular text further explains the mixing block is helpful to feature learning in complex scenarios. It is observed that ${[}L{]}_6{[}G{]}_6$ reports the best accuracy. It gives accuracy gains of 1.9\% and 6.6\% on IC13 and IC15 when compared with $none$. By placing the local mixing block in front of the global one, it is beneficial to guide the global mixing block to focus on long-term dependence capturing. On the contrary, switching their permutation is likely to confuse the role of the global mixing block, which may repetitively attend to local characteristics.

\begin{figure}[t]  
\centering  
\includegraphics[width=0.48\textwidth]{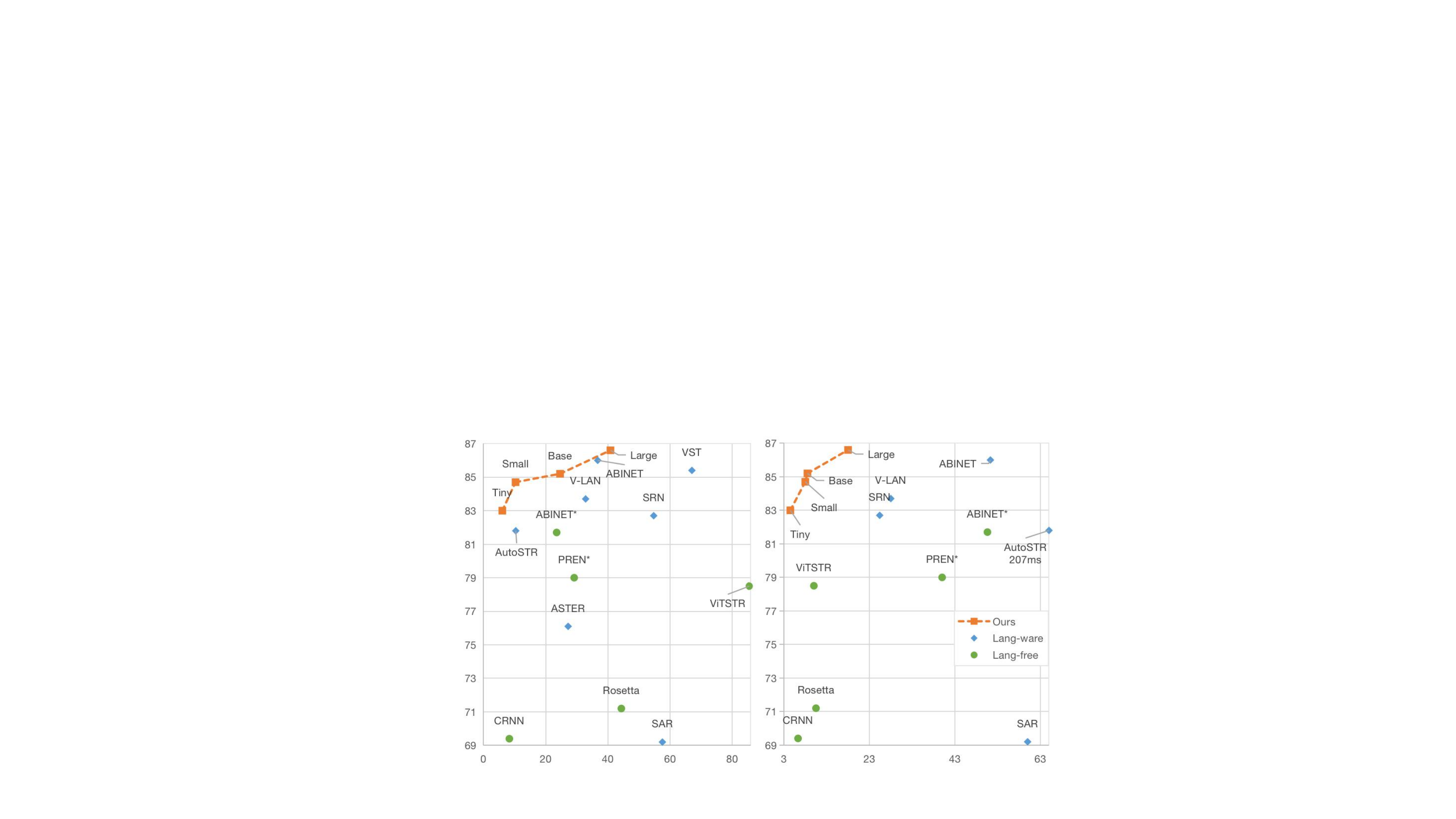}  
\caption{Accuracy-parameter (M) and Accuracy-speed (ms) plots of different models on IC15.}  
\label{fig:params_speed}  
\end{figure}

\begin{table*}[ht]
\centering
\setlength{\tabcolsep}{5pt}{
\begin{tabular}{c|r|ccc|ccc|c|cc}
\toprule
\multicolumn{2}{c|}{\multirow{2}{*}{method}} & \multicolumn{3}{c|}{English regular} & \multicolumn{3}{c|}{English unregular} & Chinese & \multirow{2}{*}{\begin{tabular}[c]{@{}c@{}}Params\\ (M)\end{tabular}} & \multirow{2}{*}{\begin{tabular}[c]{@{}c@{}}Speed\\ (ms)\end{tabular}} \\
\multicolumn{2}{c|}{}                        & IC13       & SVT      & IIIT5k      & IC15        & SVTP       & CUTE       & Scene   &                                                                       &\\
\midrule
                                & CRNN\cite{shi2017crnn}       & 91.1                                 & 81.6                                 & 82.9                                 & 69.4                                 & 70.0                                   & 65.5                                 & 53.4                                & 8.3                      & 6.3                       \\
                                & Rosetta\cite{Fedor2018Rosetta}    & 90.9                                 & 84.7                                 & 84.3                                 & 71.2                                 & 73.8                                 & 69.2                                 & -                                   & 44.3                     & 10.5                       \\
                                & SRN*\cite{yu2020srn}       & 93.2                                 & 88.1                                 & 92.3                                 & 77.5                                 & 79.4                                 & 84.7                                 & -                                   & -                        & -                       \\
                                & PREN*\cite{yan2021pren}      & 94.7                                 & 92.0                                   & 92.1                                 & 79.2                                 & 83.9                                 & 81.3                                 & -                                   & 29.1                    & 40.0                      \\
                                & ViTSTR\cite{atienza2021vitstr}     & 93.2                                 & 87.7                                 & 88.4                                 & 78.5                                 & 81.8                                 & 81.3                                 & -                                   & 85.5                    & 11.2                      \\
                                & ABINet*\cite{fang2021abinet}    & 94.9                                 & 90.4                                 & 94.6                                 & 81.7                                 & 84.2                                 & 86.5                                 & -                                   & 23.5                    & 50.6                    \\
\multirow{-7}{*}{Lan-free}     & VST*\cite{tang2021vst}       & 95.6                                 & 91.9                                 & 95.6                                 & 82.3                                 & 87.0                                 & 91.8                                 & -                                   & -                        & -                       \\
\midrule
                                & ASTER\cite{shi2019aster}      & -                                    & 89.5                                 & 93.4                                 & 76.1                                 & 78.5                                 & 79.5                                 & 54.5                                & 27.2                     & -                       \\
                                & MORAN\cite{Luo2019MORAN}      & -                                    & 88.3                                 & 91.2                                 & -                                    & 76.1                                 & 77.4                                 & 51.8                                & 28.5                     & -                       \\
                                & NRTR\cite{Sheng2019nrtr}      & 94.7                                 & 88.3                                 & 86.5                                 & -                                    & -                                    & -                                    & -                                   & 31.7                        & 160                       \\
                                & SAR\cite{li2019sar}       & 91.0                                 & 84.5                                 & 91.5                                 & 69.2                                 & 76.4                                 & 83.5                                 & 62.5                                & 57.5                     & 120                       \\
                               
                                & AutoSTR\cite{zhang2020autostr}    & -                                    & 90.9                                 & 94.7                                 & 81.8                                 & 81.7                                 & 84.0                                   & -                                   & 10.4                    & 207                     \\
                                & SRN\cite{yu2020srn}        & 95.5                                 & 91.5                                 & 94.8                                 & 82.7                                 & 85.1                                 & 87.8                                 & 60.1                                & 54.7                       & 25.4                       \\
                                & PREN2D\cite{yan2021pren}     & 96.4                                 & 94.0                                   & 95.6                                 & 83.0                                   & 87.6                                 & 91.7                                 & -                                   & -                        & -                       \\
                                & VisionLAN\cite{Wang_2021_visionlan}      & 95.7                                 & 91.7                                 & 95.8                                 & 83.7                                 & 86.0                                   & 88.5                                 & -                                   & 32.8                    & 28.0                      \\
                                & ABINet\cite{fang2021abinet}     & {\textbf{97.4}} & {93.5}                           & 96.2                                 & {86}                             & {89.3}                           & 89.2                                 & -                                   & 36.7                     & 51.3                    \\
\multirow{-11}{*}{Lan-aware}    & VST\cite{tang2021vst}        & 96.4                                 & {\textbf{93.8}} & {\textbf{96.3}} & 85.4                                 & 88.7                                 & {\textbf{95.1}} & \textbf{-}                          & 64.0                    & -                       \\
\midrule
                                & \multicolumn{1}{c|}{SVTR-T (Tiny)}     & 96.3                                & 91.6                                 & 94.4                                 & 84.1                                 & 85.4                                 & 88.2                                 & 67.9                                   & 6.03                     & 4.5                    

\\
                                
                                & \multicolumn{1}{c|}{SVTR-S (Small)}     & 95.7                                & 93.0                                   & 95.0                                   & 84.7                                   & 87.9                                   & 92.0                                   & 69.0                                   & 10.3                                  & 8.0                       \\
                            & \multicolumn{1}{c|}{SVTR-B (Base)}     & 97.1                                 & 91.5                                 & 96.0                                   & 85.2                            &\textbf{89.9}                               & 91.7                                 & 71.4                                   & 24.6                    &     8.5                     \\
\multirow{-4}{*}{Ours}          

                            & \multicolumn{1}{c|}{SVTR-L (Large)}     & {97.2}                           & 91.7                                 &  \textbf{96.3} & \textbf{86.6} & 88.4                                 & \textbf{95.1} & {\textbf{72.1}} & 40.8                    & 18.0                      \\
\bottomrule
\end{tabular}}
\caption{Results on six English and one Chinese benchmarks tested against existing methods, where $CRNN$ and $Rosetta$ are from the reproduction of CombBest \protect\cite{Baek2019WhatIW}. $Lan$ means language and * means the language-free version of the corresponding method. The speed is the inference time on one NVIDIA 1080Ti GPU averaged over 3000 English image text.}
\label{tab:sota}
\end{table*}

\subsection{Comparison with State-of-the-Art}


We compare SVTR with previous works on six English benchmarks covering regular and irregular text and one Chinese scene dataset in Table \ref{tab:sota}. The methods are grouped by whether utilizing the language information, i.e., lan-free and lan-aware. We first look into the result in English datasets. Even SVTR-T, the smallest one of the SVTR family, achieves highly competitive accuracy among the lan-free ones. While other SVTRs are new state-of-the-art on most datasets. When further compared with lan-aware ones, SVTR-L gains the best accuracy on IIIT, IC15 and CUTE among the six English benchmarks. Its overall accuracy is on par with recent studies \cite{fang2021abinet,tang2021vst}, which used extra language models. As a contrast, SVTR enjoys its simplicity and runs faster. The results basically imply that a single visual model also could perform the recognition well, as discriminative character features are successfully extracted.

We then analyze the result on Chinese Scene Dataset. \cite{chen2021benchmarking} gives the accuracy of six existing methods as shown in the table. Encouragingly, SVTR performs considerably well. Compared with SAR, the best one among the listed ones, accuracy gains ranging from 5.4\% to 9.6\% are observed, which are noticeable improvements. The result is explained as SVTR comprehensively perceives multi-grained character component features. They are well suited to characterize Chinese words that have rich stroke patterns. 

In Figure.\ref{fig:params_speed}, we also depict the accuracy, parameter and inference speed of different models on IC15. Owning to their simpleness, SVTRs consistently rank top-tier in both accuracy-parameter and accuracy-speed plots, further demonstrating its superiority compared to existing methods.

\subsection{Visualization Analysis}

\begin{figure}[ht]  
\centering  
\includegraphics[width=0.48\textwidth]{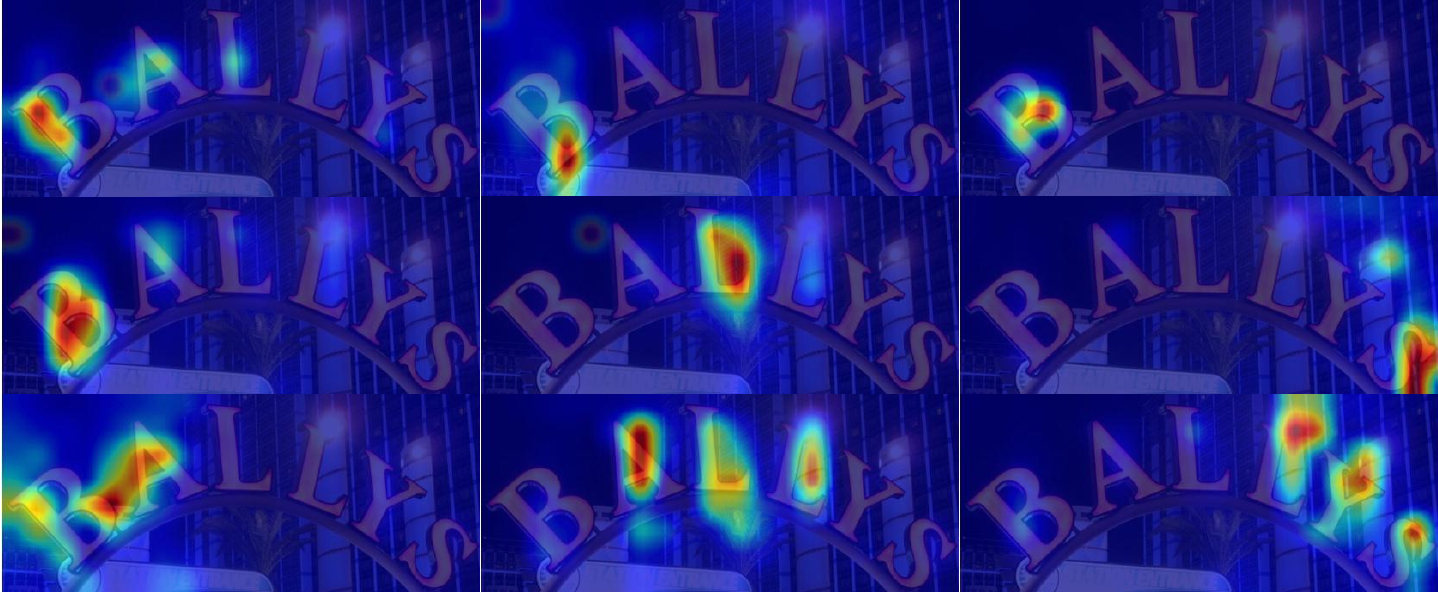}  
\caption{Visualization of SVTR-T attention maps.}  
\label{fig:vis}  
\end{figure}

We have visualized the attention maps of SVTR-T when decoding different character components. Each map can be explained as serving a different role in the whole recognition. Nine maps are selected for illustration, as shown in Figure.\ref{fig:vis}. The first line shows three maps of gazing into a fraction of character "B", with emphasis on its left side, bottom and middle parts, respectively. It indicates that different character regions are contributed to the recognition. The second line exhibits three maps of gazing into different characters, i.e., "B", "L", and "S". SVTR-T is also able to learn its feature by viewing a character as a whole. While the third line exhibits three maps simultaneously activate multiple characters. It implies that dependencies among different characters are successfully captured. The three lines together reveal that sub-character, character-level and cross-character clues are all captured by the recognizer, in accordance with the claim that SVTR perceives multi-grained character component features. It again explains the effectiveness of SVTR.

With the results and discussion, we conclude with that SVTR-L enjoys the merits of accurate, fast and versatile, being a highly competitive choice in accuracy-oriented applications. While SVTR-T is an effective and much smaller model yet quite fast. It is appealing in resource-limited scenarios.

\section{Conclusion}
We have presented SVTR, a customized visual model for scene text recognition. It extracts multi-grained character features that describe both stroke-like local patterns and inter-component dependence of varying distance at multiple height scales. Therefore, the recognition task can be conducted by using a single visual model, enjoying the merits of accuracy, efficiency and cross-lingual versatility. SVTR with different capacities are also devised to meet diverse application needs. Experiments on both English and Chinese benchmarks basically verify the proposed SVTR. Highly competitive or even better accuracy is observed compared to state-of-the-art methods, while running faster. We hope that SVTR will foster further research in scene text recognition.
\section*{Acknowledgments}

The work is supported by the National Nature Foundation of China (No. 62172103, 61876016) and CCF-Baidu Open Fund (No.2021PP15002000). 

\bibliographystyle{named}
\bibliography{ijcai22}

\end{document}